 \documentclass[runningheads]{llncs}
\usepackage[T1]{fontenc}

\usepackage{amssymb}

\usepackage{amsmath}
\usepackage{booktabs}
\usepackage{tabularray}
\usepackage{amssymb}
\usepackage{amsmath}
\usepackage{subcaption}
\usepackage{graphicx}
\usepackage{multirow}
\usepackage{multicol}
\usepackage{url}
\usepackage{hyperref}

\usepackage{breakurl}
\usepackage{comment}
\usepackage[none]{hyphenat}
\usepackage{adjustbox}
\usepackage{microtype}
\usepackage{ragged2e}
\usepackage{xcolor}
\usepackage{caption}

\begin{document}
\title{Integrating Weather Station Data and Radar for Precipitation Nowcasting: SmaAt-fUsion and SmaAt-Krige-GNet}

\titlerunning{SmaAt-fUsion and SmaAt-Krige-GNet}
%
%

\author{Jie Shi \and Aleksej Cornelissen \and Siamak Mehrkanoon}

\institute{
Department of Information and Computing Sciences, 
Utrecht University, Utrecht, The Netherlands \\
\email{j.shi1@uu.nl, aleksej.cornelissen@hotmail.com, s.mehrkanoon@uu.nl}
}

\maketitle              


\begin{abstract}
Short-term precipitation nowcasting is essential for flood management, transportation, energy system operations, and emergency response. However, many existing models fail to fully exploit the extensive atmospheric information available, relying primarily on precipitation data alone. This study examines whether integrating multi variable weather-station measurements with radar can enhance nowcasting skill and introduces two complementary architectures that integrate multi variable station data with radar images. The SmaAt-fUsion model extends the SmaAt-UNet framework by incorporating weather station data through a convolutional layer, integrating it into the bottleneck of the network; The SmaAt-Krige-GNet model combines precipitation maps with weather station data processed using Kriging, a geo-statistical interpolation method, to generate variable-specific maps. These maps are then utilized in a dual-encoder architecture based on SmaAt-GNet, allowing multi-level data integration. Experimental evaluations were conducted using four years (2016--2019) of weather station and precipitation radar data from the Netherlands. Results demonstrate that SmaAt-Krige-GNet outperforms the standard SmaAt-UNet, which relies solely on precipitation radar data, in low precipitation scenarios, while SmaAt-fUsion surpasses SmaAt-UNet in both low and high precipitation scenarios. This highlights the potential of incorporating discrete weather station data to enhance the performance of deep learning-based weather nowcasting models.

\end{abstract}


\section{Introduction}

Precipitation nowcasting refers to the short-term forecasting of rainfall intensity, typically within a timeframe ranging from a few minutes to several hours. Accurate nowcasting is crucial for various weather-dependent sectors, including flood management, transportation, energy distribution, and emergency response \cite{li2024quantitative,wilson2010nowcasting}. However, although numerical weather prediction (NWP) and radar-based methods have made notable progress, there remains improvement in operational settings. NWP models utilize physical atmospheric properties to generate multiple precipitation scenarios, but they are computationally expensive and often impractical for short-term forecasts due to their sensitivity to initial conditions and require high computations \cite{bauer2015quiet,wilks2011statistical}. This gap poses a growing threat to the effectiveness of early warning systems, particularly in densely populated or infrastructure-critical regions such as the Netherlands. The effectiveness of nowcasting models depends on their ability to capture the complex, dynamic nature of precipitation patterns across different spatial and temporal scales. 

Recent advancements in deep learning have led to the development of advanced data-driven models that utilize historical weather data to capture complex spatiotemporal patterns \cite{mehrkanoon2019deep,vatamany2025graph,ham2023anthropogenic,allen2025end}. Convolutional Neural Networks (CNNs) based models, in particular, have demonstrated promising results in precipitation nowcasting by effectively extracting spatial features from radar images \cite{ayzel2019all,ayzel2020rainnet,jiang2024hybrid}. However, these models almost exclusively rely on radar reflectivity data, overlooking other atmospheric drivers, such as temperature, humidity, pressure, and wind, which fundamentally influence precipitation formation. As a result, they are unable to capture the localized signals that often precede rainfall initiation. Consequently, radar-only systems may fail to capture pre-convective signals, limiting both accuracy and generalization under rapidly changing weather conditions.


Some recent studies have attempted to incorporate multiple atmospheric variables into their models to improve prediction robustness \cite{wang2025enhancing,cao2025enhancing}. For instance, the authors in \cite{kaparakis2023wf} introduced WF-UNet, a two-stream UNet architecture that processes both precipitation maps and wind speed maps to assess the impact of wind on precipitation nowcasting. However, such approaches typically depend on spatially continuous maps rather than integrating discrete station-based observations. Weather stations provide highly localized, multi-variable atmospheric observations that offer valuable insights into the physical processes behind rainfall. However, station data alone lacks spatial continuity, making it challenging to use directly in deep learning models designed for grid-based data. Our work aims to bridge this gap by learning from both discrete weather station data and radar-based precipitation maps.


To address this limitation, this study explores the integration of additional meteorological observations into deep learning-based precipitation nowcasting. We propose two architectures, SmaAt-fUsion and SmaAt-Krige-GNet, that incorporate multi-variable weather station data together with radar observations. The main contributions are summarized as follows.

\begin{enumerate}
    \item \textbf{Multi-source data integration.} We combine four years (2016–2019) of radar precipitation data with multi-variable observations from 22 KNMI weather stations in the Netherlands, enabling joint learning from radar grids and station measurements.

    \item \textbf{SmaAt-fUsion.} We extend the SmaAt-UNet framework \cite{trebing2021smaat} by injecting weather station information at the bottleneck layer to enhance high-level feature representations.

    \item \textbf{SmaAt-Krige-GNet.} We generate spatially continuous maps from station observations using Kriging interpolation and incorporate them into SmaAt-UNet model by incorporating a dual-encoder architecture, ensuring that the model learns from both data sources at various levels of abstraction.
\end{enumerate}

\section{Methodology}
\subsection{Dataset}

This study utilizes two complementary datasets: a radar-based precipitation dataset and a weather station dataset. Both datasets were obtained from the Royal Netherlands Meteorological Institute (KNMI \footnote{ \url{https://www.knmi.nl/home.}}) and are used together to improve precipitation nowcasting performance.

\noindent \textbf{Precipitation Dataset:}
The precipitation dataset, originally used by Trebing et al. \cite{trebing2021smaat}, consists of radar-based rainfall data collected over four years (2016–2019) in the Netherlands. It contains approximately 420,000 rain maps, each captured at 5-minute intervals. The data is sourced from two C-band Doppler weather radar stations located in:
De Bilt (52.10°N, 5.18°E, 44 m MSL) and Den Helder (52.96°N, 4.79°E, 51 m MSL). To generate rain maps, each radar performs four full 360° azimuthal scans at beam elevation angles of 0.3°, 1.1°, 2.0°, and 3.0° \cite{precipitation_dataset,trebing2021smaat}.

\begin{figure}[!ht]
    \centering
    \frame{\includegraphics[width=0.35\linewidth]{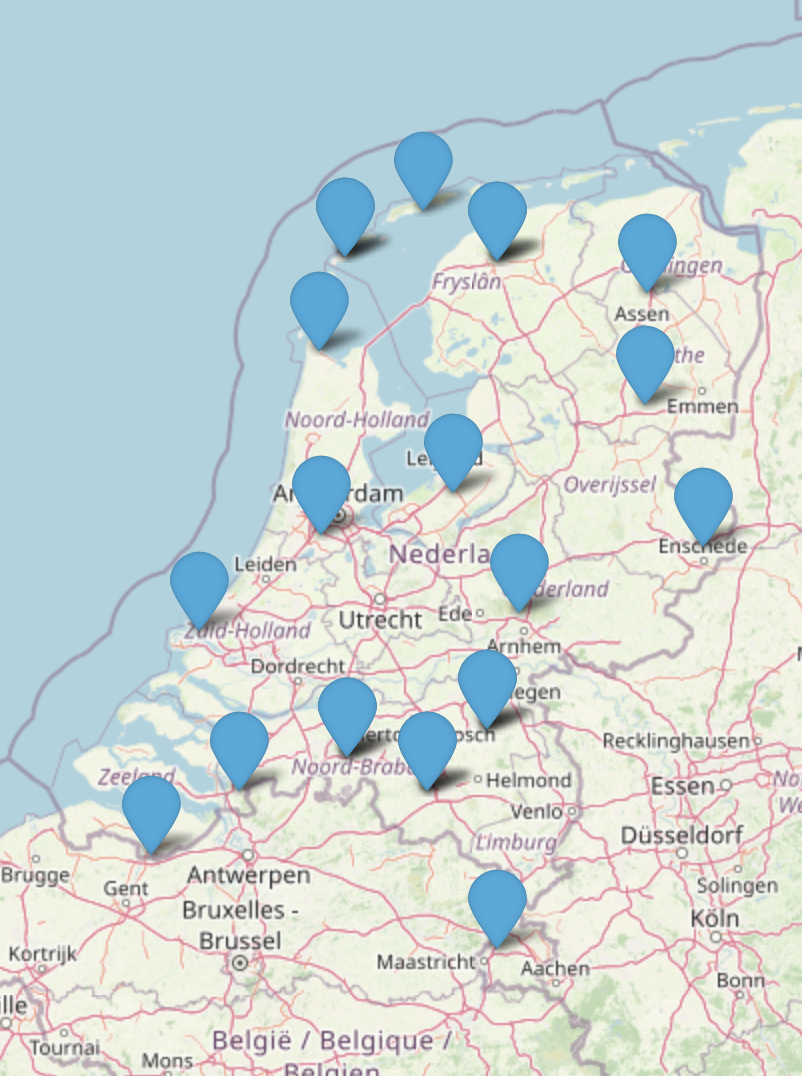}}
    \captionsetup{font=scriptsize}
    \caption{An overview map of all the nodes included in the weather station dataset. The dataset contains a total of 22 weather stations. Some stations share identical coordinates, resulting in 17 nodes in the figure.}
    \label{fig:stations_map}
\end{figure}

\noindent \textbf{Weather Station Dataset:}
The second dataset consists of weather station data, which provides multivariable atmospheric measurements from 22 weather stations across the Netherlands (see Fig. \ref{fig:stations_map}). This dataset is novel and contains 8 meteorological variables, selected based on availability, relevance, and consistency. The recorded variables are: Temperature, Humidity, Atmospheric Pressure, Wind Speed, Max Wind Speed, Wind Speed Deviation, Wind Direction, Wind Direction Deviation.

\noindent \textbf{Dataset Integration and Processing:}
To create a unified dataset for model input, the precipitation and weather station datasets undergo several preprocessing steps to ensure consistency and alignment. For the precipitation maps, a 288×288 pixel cutout is taken from the center of the Netherlands, covering the area with the highest data availability. The selected geographic boundaries are: Longitude: 4.29°E to 6.73°E and
Latitude: 51.32°N to 52.81°N. These maps are then resized to 64×64 pixels to reduce memory requirements, as both datasets are incorporated into the models. Finally, the precipitation values are normalized using the maximum precipitation value in the dataset (47.83 mm). The precipitation dataset contains a significant number of radar images with little to no recorded rainfall. To mitigate the risk of biasing the model toward predicting zero values, precipitation maps where fewer than 50\% of the pixels indicate rainfall are excluded. Once filtered, the remaining precipitation maps are temporally matched with the corresponding weather station data recorded at the same timestamp. Any data point that does not have a corresponding entry in the other dataset is removed, ensuring that only fully aligned data pairs are retained. To maintain consistency across variables, all weather station measurements are standardized based on their respective means and standard deviations. A challenge in merging the two datasets arises from their differing temporal resolutions. While precipitation maps are recorded at 5-minute intervals, weather station data is available at 10-minute intervals. To resolve this mismatch, the weather station dataset is duplicated to align with the precipitation timestamps, allowing for seamless integration of both sources. After preprocessing, the final dataset is divided into a training set (2016–2018) and a testing set (2019). During model training, a validation set is created by randomly selecting 10\% of the training data.

\begin{figure*}[tph]
    \centering
    \includegraphics[width=1\linewidth]{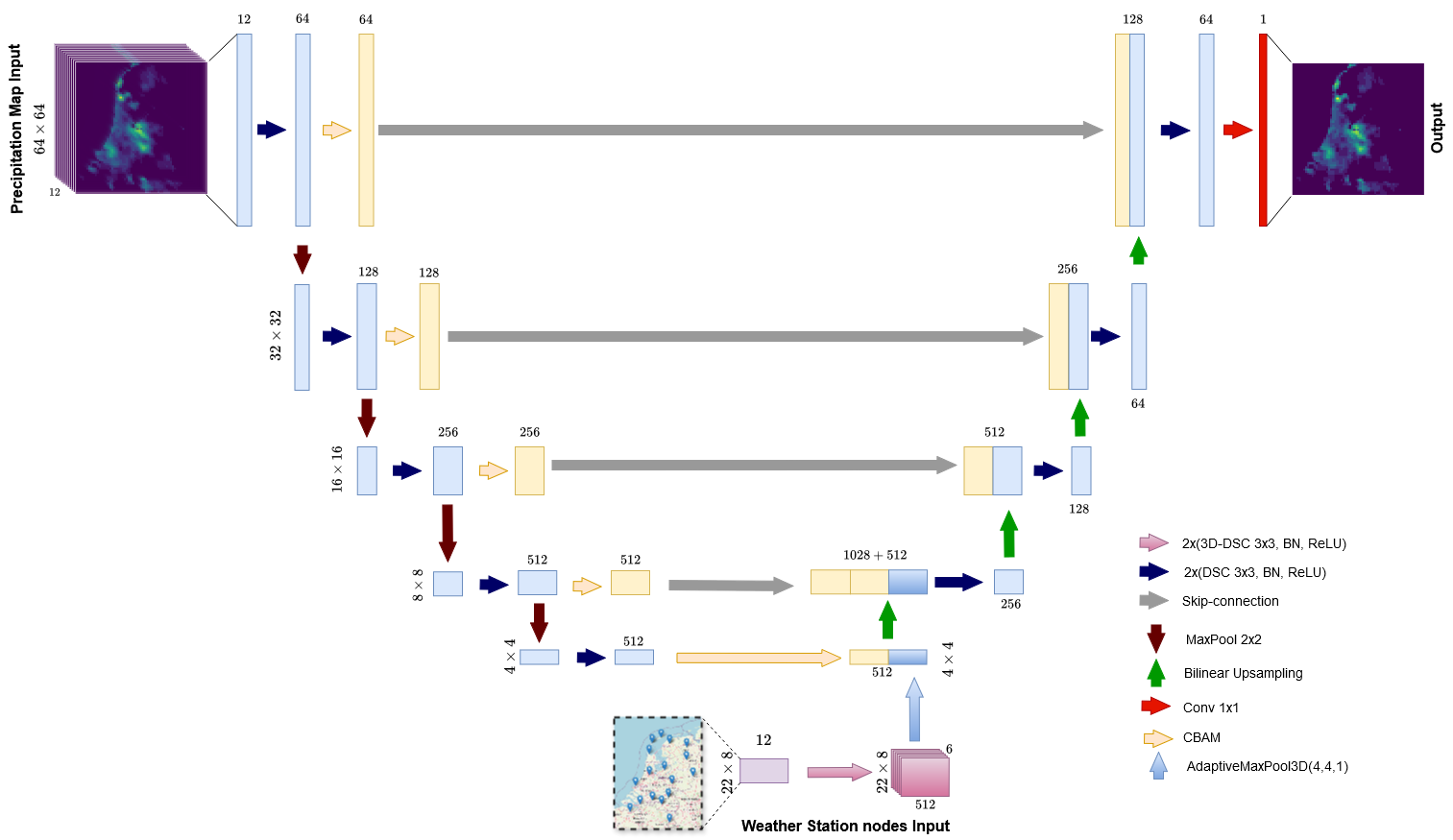}
    \captionsetup{font=scriptsize}
    \caption{Overview of the SmaAt-fUsion architecture. Weather station data is integrated into the model by concatenating its learned representation with that of the precipitation maps at the bottleneck of the SmaAt-UNet model.}
    \label{fig:smaat_fusion}
\end{figure*}

\begin{table}
    \centering
        \caption{Size of each train and test set for the 4-year precipitation dataset}
    \begin{adjustbox}{max width=\linewidth}
    \begin{tabular}{l c c c}

        \toprule
         \textbf{Rain Threshold} & \textbf{Train Size} & \textbf{Test Size} & \textbf{Subset} \\
         \midrule
          $0\%$ (Original) &  314940 & 105003 & 100\%  \\
         $50\%$ & 5734 & 1557 & 1.74\% \\
         \bottomrule
    \end{tabular}
    \end{adjustbox}
    \label{tab:my_label}
\end{table}

\subsection{Proposed models}

In the following, we introduce two architectures designed to extend the core SmaAt-UNet model \cite{trebing2021smaat} to learn from both weather station data and precipitation maps. The key difference between these models lies in how they integrate weather station data into SmaAt-UNet, which primarily operates on gridded precipitation maps. Both models take 12 precipitation maps as input, representing one hour of weather radar data. In addition to these maps, they incorporate station data using different integration strategies. The output of each model is a single precipitation map predicting conditions 30 minutes after the last input frame.

\begin{figure*} 
    \centering
    \begin{subfigure}[]{1\textwidth}
        \centering
        \includegraphics[width=\linewidth]{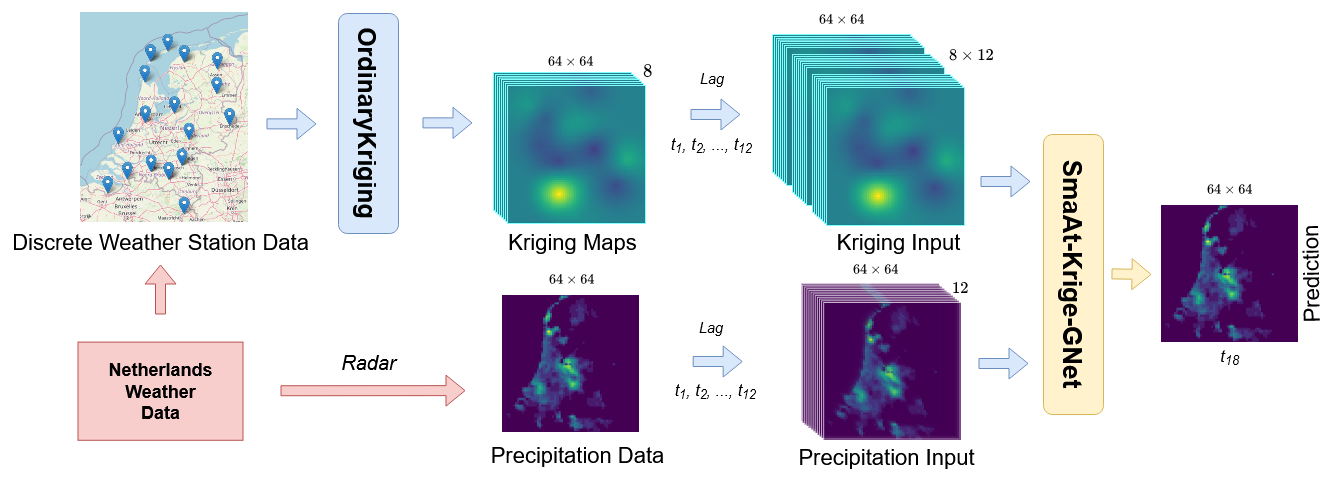}
        \captionsetup{font=scriptsize}
        \caption{Overview of the Kriging process. For each variable, a Kriging map is generated for each timestep. Subsequently, 12 timesteps are fed into the SmaAt-Krige-GNet model.}
        \label{fig:gnet_overview}
    \end{subfigure}
    \vspace{1em} 
    \begin{subfigure}[]{1\textwidth}
        \centering
        \includegraphics[width=\linewidth]{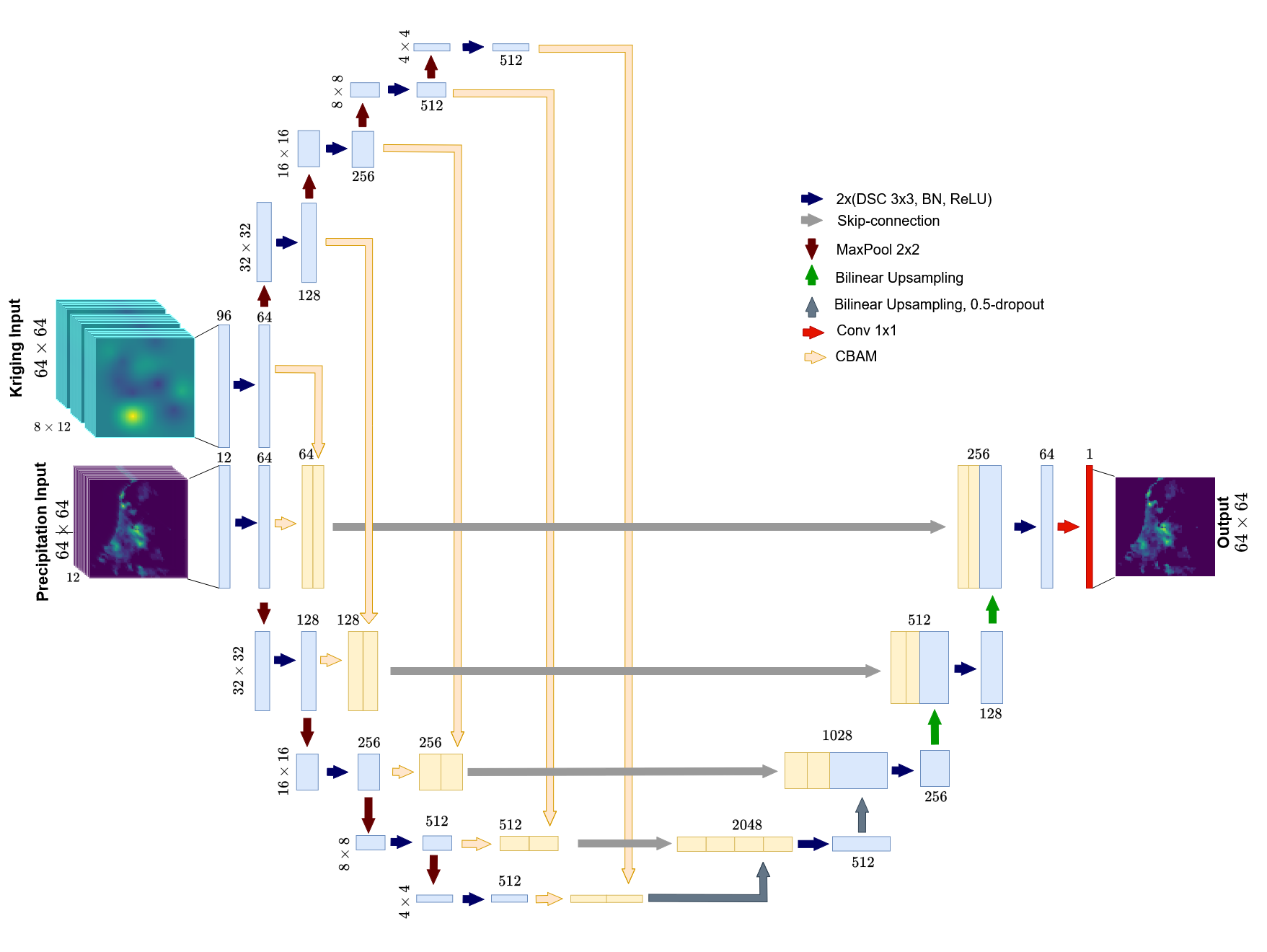}
        \captionsetup{font=scriptsize}
        \caption{The SmaAt-Krige-GNet model. The upper stream processes Kriging data and mirrors the layers of the bottom stream, which handles precipitation data. At each level, the output of the upper stream is concatenated channel-wise with the corresponding output of the bottom stream.}
        \label{fig:gnet}
    \end{subfigure}
    \captionsetup{font=scriptsize}
    \caption{An illustration of the Kriging process and Overview of the SmaAt-Krige-GNet architecture.}
    \label{fig:combined_overview}
\end{figure*}

\subsection{SmaAt-fUsion model}

In SmaAt-fUsion, our aim is to integrate weather station data into the core SmaAt-Unet model \cite{trebing2021smaat}, which operates on gridded precipitation radar maps. To this end, following the approach of \cite{mehrkanoon2019deep}, we exploit the spatio-temporal structure of station data by transforming each regressor vector, containing historical weather variable measurements from all stations, into a tensor with dimensions (stations = 22, variables = 8, lags = 12). This tensor is processed through a double 3D depthwise separable convolution block, consisting of a depthwise convolution followed by a pointwise convolution. Compared to classical 3D convolutions, this decomposition significantly reduces the number of parameters while preserving spatial and temporal correlations in the data. The output from this convolutional block is then passed through a 3D Adaptive Max Pooling layer, which reshapes it to match the bottleneck tensor size of the original SmaAt-Unet model. Finally, the pooled representation is concatenated with the bottleneck tensor, as illustrated in Fig. \ref{fig:smaat_fusion}. By integrating station data, we enrich the model's latent space with additional insights from weather station observations. This fusion strengthens the model’s ability to better capture complex spatio-temporal dependencies, ultimately improving its performance in precipitation mapping.

\subsection{SmaAt-Krige-GNet model}

In this work, we explicitly integrate spatial dependencies between weather stations into the SmaAt-UNet model. To achieve this, we first generate grid-based maps of station variables using Kriging \cite{Isaaks1989,Matheron1963}. Inspired by \cite{reulen2024ga}, which introduced SmaAt-GNet as a generator within an adversarial learning framework for precipitation nowcasting, we propose SmaAt-Krige-GNet. This model combines the grid-based maps derived from station data with precipitation maps. The key distinction between SmaAt-GNet and the standard SmaAt-UNet model lies in the addition of an extra encoder stream. This new stream mirrors the encoder layers of SmaAt-UNet and concatenates its representations at each level with those of the original SmaAt-UNet encoder. While the original SmaAt-GNet in \cite{reulen2024ga} was designed to integrate binary precipitation masks into the SmaAt-UNet model, our proposed SmaAt-Krige-GNet instead utilizes Kriging-based spatial maps. By structuring the model into two independent streams rather than stacking all inputs channel-wise in a single-stream architecture, SmaAt-Krige-GNet can independently learn hidden features from both weather and precipitation data. Since all models process 12 input images, we modify the secondary input of SmaAt-GNet to also accept 12 images. Given that there are 8 weather variables, the resulting input forms a 4D tensor of size $64\times64\times8\times12$, which is then flattened into a $64\times64\times96$ layer, as illustrated in Fig. \ref{fig:gnet}. To explicitly incorporate spatial data, Kriging is applied to interpolate weather station measurements into 2D geospatial maps, ensuring that each weather variable at each timestep is represented in the same spatial resolution as the precipitation maps.

\section{Experimental setup and evaluation}
\subsection{Model setup}
The models were implemented and trained using PyTorch Lightning. The training process used Adam Optimizer with a learning rate of 0.001. To avoid overfitting, early stopping is applied with a maximum patience of $ES = 12$ to monitor the loss of validation. In addition, we apply learning rate scheduling with a patience of $LR = 8$ epochs, which reduces the learning rate when no improvement in validation loss was observed. The models were trained on the dataset for a maximum of 200 epochs, with a batch size of 16. These hyperparameters were chosen based on preliminary experimentation to balance model performance and computational efficiency. 


To generate the Kriging maps for SmaAt-Krige-GNet, we use Ordinary Kriging from PyKrige \cite{pykrige}. A grid is defined with the same resolution as the precipitation maps, specifically $64 \times 64$. Subsequently, a variogram is fitted using the spherical model, which has been shown to be suitable for meteorological applications \cite{Verworn2011,CSEGRecorderVariograms}. To ensure accurate spatial representation, we specify the coordinate type as geographic, allowing the model to directly use the latitude and longitude of each weather station. However, Kriging with PyKrige can be numerically unstable when using its default inverse matrix algorithm. To mitigate this, we set the `pseudo\_inv' parameter of Ordinary Kriging to True, which stabilizes computations at the cost of increased processing time \cite{pykrige}. Once the linear system is solved, a set of weights $\lambda_i$ is obtained, which are then applied to interpolate the Kriging value at each coordinate within the grid. This process generates spatially continuous weather variable maps that align with the resolution of the precipitation data.

\begin{figure}
    \centering
    \includegraphics[width=0.85\linewidth]{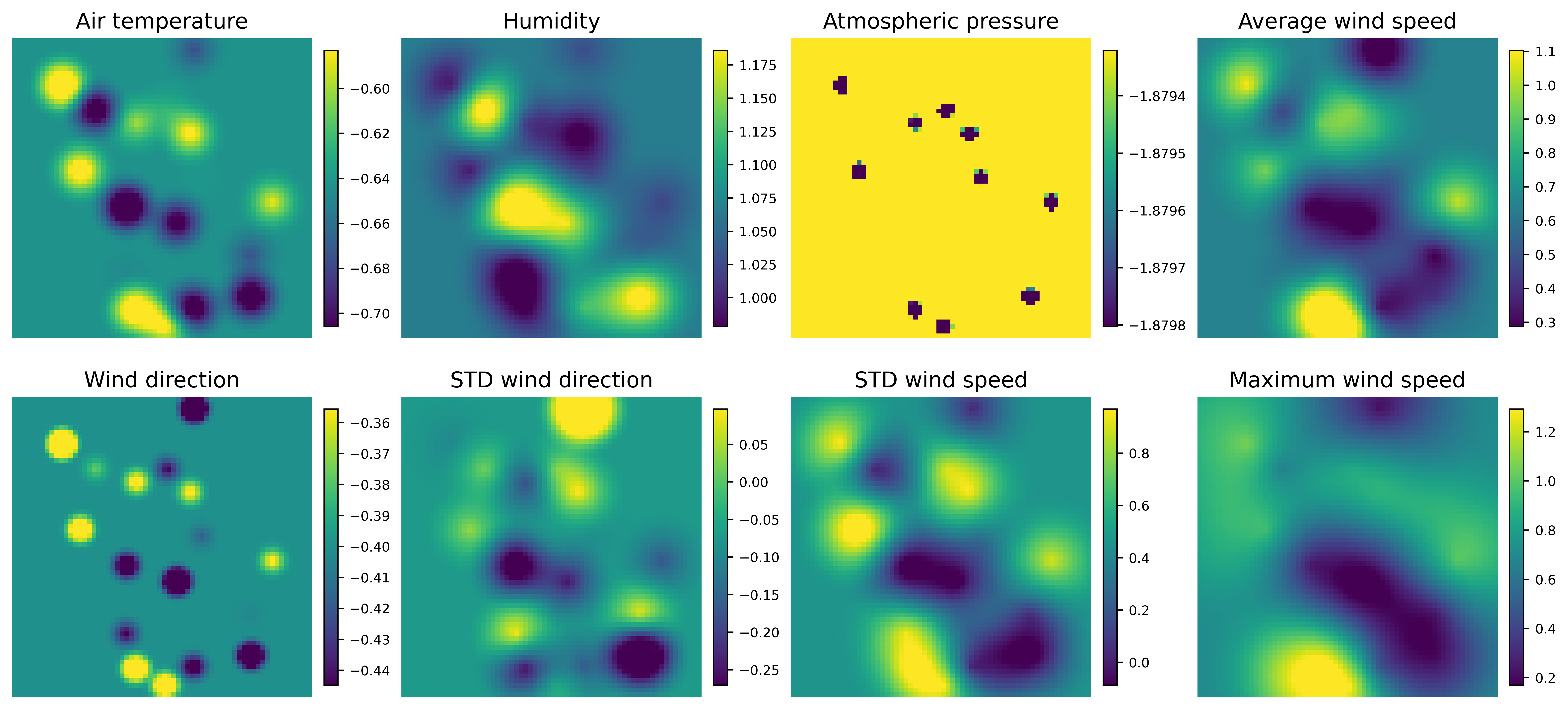}
    \captionsetup{font=scriptsize}
    \caption{An example of Kriging map input of the eight meteorological variables at a selected timestamp. To improve visibility, Atmospheric pressure uses a percentile-based color scale due to its narrow dynamic range, while the other variables share a common normalization ($-2$ to 2).}
    \label{fig:kriging_example}
\end{figure}



\subsection{Evaluation}
The models are trained using the Mean Squared Error (MSE) loss between predicted and ground truth precipitation maps. Following previous studies \cite{reulen2024ga,trebing2021smaat,stanczyk2021deep}, model performance is evaluated using several binary precipitation nowcasting metrics, including the F1-score, Critical Success Index (CSI), Heidke Skill Score (HSS), and Matthews Correlation Coefficient (MCC). Since precipitation maps contain continuous values, predictions are thresholded to generate binary rain masks. We evaluate three commonly used rainfall thresholds: 0.5 mm/h (light precipitation), 10 mm/h, and 20 mm/h, enabling assessment across different precipitation intensities. The metrics are computed based on true positives (TP), false positives (FP), true negatives (TN), and false negatives (FN). In particular, CSI, HSS, and MCC are defined as follows:
\begin{equation}
    \begin{aligned}
 \small
        \textbf{CSI}         &= \frac{\text{TP}}{\text{TP} + \text{FP} + \text{FN}}, \\
        \textbf{HSS}         &= \frac{2 \cdot (\text{TP} \cdot \text{TN} - \text{FP} \cdot \text{FN})}{(\text{TP} + \text{FN}) \cdot (\text{FN} + \text{TN}) + (\text{TP} + \text{FP}) \cdot (\text{FP} + \text{TN})}, \\
        \textbf{MCC}         &= \frac{\text{TP} \cdot \text{TN} - \text{FP} \cdot \text{FN}}{\sqrt{(\text{TP} + \text{FP})(\text{TP} + \text{FN})(\text{TN} + \text{FP})(\text{TN} + \text{FN})}}.
    \end{aligned}
\end{equation}

\begin{figure*}
    \centering
    \includegraphics[width=0.8\linewidth]{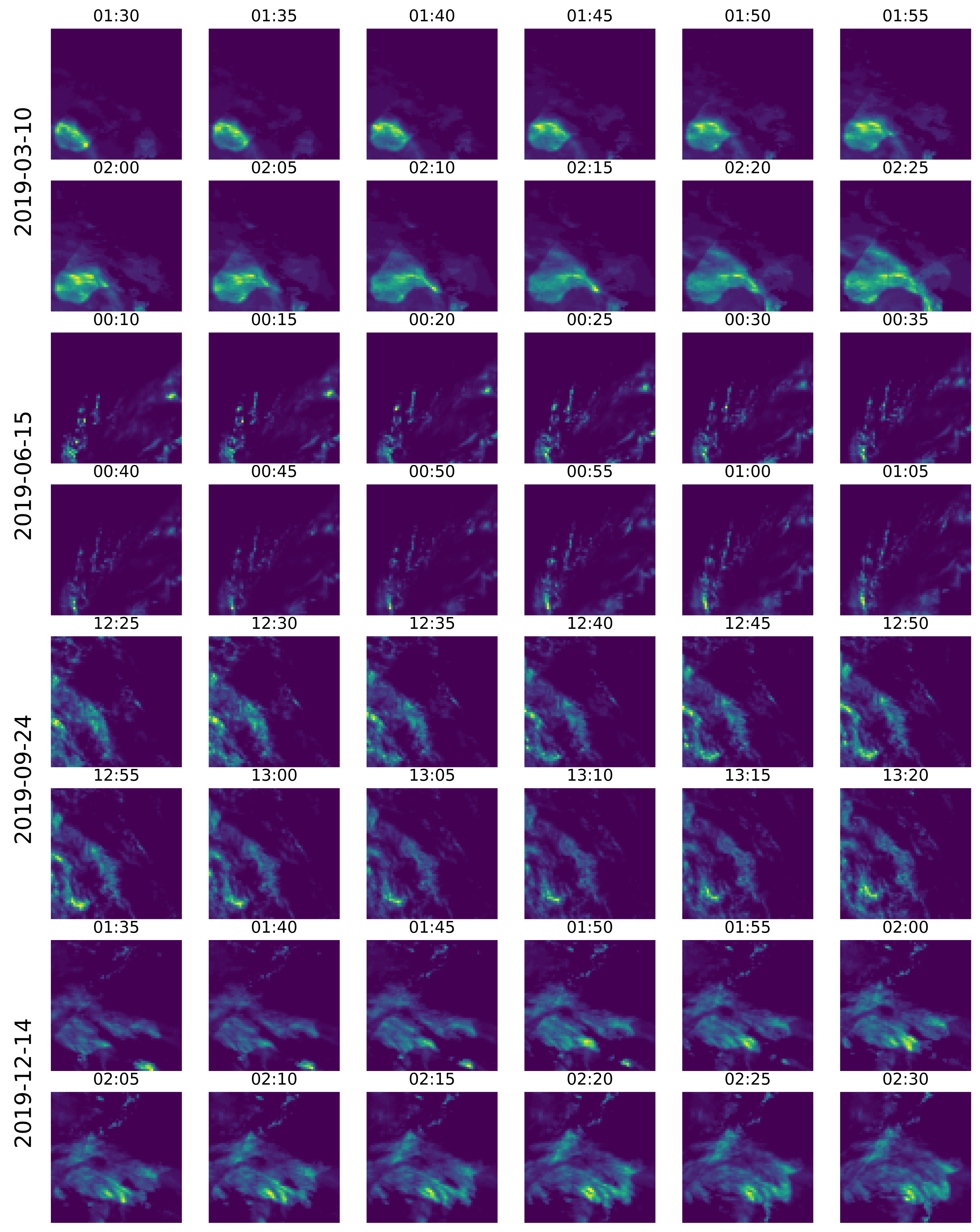}
    \captionsetup{font=scriptsize}
    \caption{Examples precipitation input of the dataset on 2019-03-10, 2019-06-15, 2019-09-29, 2019-12-14.}
    \label{fig:input12}
\end{figure*}

\begin{figure*}
    \centering
    \includegraphics[width=0.85\linewidth]{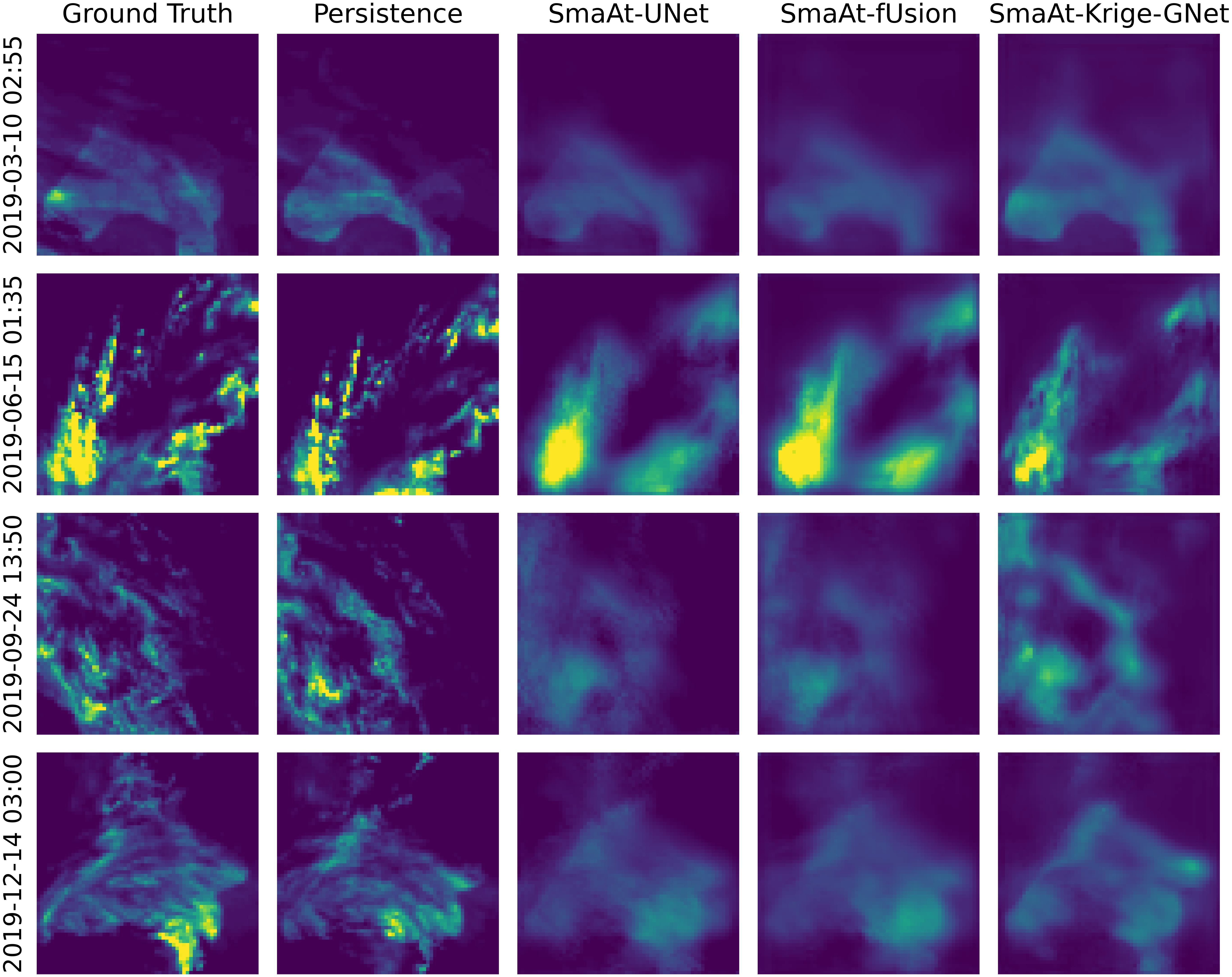}
    \captionsetup{font=scriptsize}
    \caption{Example outputs of each model, using the input seen in Fig. \ref{fig:input12}} 
    \label{fig:example-outputs}
\end{figure*}

\begin{table*}
\centering
\scriptsize
\caption{Model performance metrics across different precipitation thresholds for the studied dataset.}
\begin{tabular}{c c c c c c c}
\toprule
\textbf{Threshold} & \textbf{Model} & \textbf{MSE (pixel) $\downarrow$}  & \textbf{F1 Score $\uparrow$} & \textbf{CSI $\uparrow$} & \textbf{HSS $\uparrow$} & \textbf{MCC $\uparrow$} \\
\midrule
$\geq 0.5 \text{ mm/h}$ 
  & Persistence         & 0.065986                      & 0.678436          & 0.513359          & 0.241484          & 0.483398          \\
                     & SmaAt\_UNet & 0.043293& 0.763476 &0.617437& 0.299234 &0.601438\\
                     & SmaAt-fUsion & 0.035778 & 0.768003 & 0.623381 & 0.297824 & 0.604511 \\
                     & SmaAt-Krige-GNet & \textbf{0.035618} & \textbf{0.780774} & \textbf{0.640385} & \textbf{0.315630} & \textbf{0.633025} \\
                     
\midrule
$\geq 10 \text{ mm/h}$     
    & Persistence         & -                                   & 0.117185          & 0.062239          & 0.056982          & 0.113965          \\
                     & SmaAt\_UNet& -& 0.144665& 0.077972& 0.071221& 0.148874 \\
                     & SmaAt-fUsion & - & \textbf{0.187187} & \textbf{0.103258} & \textbf{0.092433} & \textbf{0.189564} \\
                     & SmaAt-Krige-GNet & - & 0.052808 & 0.027120 & 0.026092 & 0.091192 \\
                     
\midrule
$\geq 20 \text{ mm/h}$     
    & Persistence         & -                                    & 0.033757          & 0.017169          & 0.016623          & 0.033265          \\
                     & SmaAt\_UNet & -& 0.047089& 0.024113 & 0.023404 & 0.051688  \\
                     & SmaAt-fUsion & - & \textbf{0.052607} & \textbf{0.027014} & \textbf{0.026190} & \textbf{0.061962} \\
                     & SmaAt-Krige-GNet & - & 0.029783 & 0.015117 & 0.014851 & 0.054041 \\
                    
\bottomrule
\end{tabular}
\label{tab:performance}
\end{table*}

\section{Results and Discussion}

\subsection{Model performance}
After training, models were selected based on the lowest validation loss to ensure optimal generalization. The obtained results are presented in Table \ref{tab:performance}. These results were computed after denormalizing the data, following the approach used in similar studies \cite{reulen2024ga,trebing2021smaat}.

The results in Table \ref{tab:performance} show that the proposed SmaAt-fUsion model outperforms the classical SmaAt-UNet across all tested thresholds. This suggests that incorporating station data effectively guides the SmaAt-UNet core model in better learning atmospheric dynamics. Furthermore, the SmaAt-fUsion model achieves superior performance compared to SmaAt-Krige-GNet at higher rainfall thresholds of 10 mm/h and 20 mm/h. However, for low-intensity rainfall, SmaAt-Krige-GNet performs best among the examined models. This indicates that integrating additional weather variables at multiple levels of the SmaAt-UNet encoder enables the model to extract more detailed information from the augmented weather station data, compared to the SmaAt-fUsion model.

At higher rainfall thresholds (10 mm/h and 20 mm/h), SmaAt-Krige-GNet underperforms relative to both SmaAt-fUsion and the core SmaAt-UNet in terms of binary evaluation metrics. This performance decrease is likely due to the model’s tendency to over-smooth predictions, primarily a consequence of incorporating Kriging-generated maps. As shown in Fig. \ref{fig:kriging_example}, these maps inherently introduce smoothness. Although Kriging provides spatial continuity required by convolutional architectures, the limited number of station observations reduces its ability to reconstruct fine-scale atmospheric structures. This smoothing effect is especially noticeable in variables such as atmospheric pressure and wind-related fields, where local gradients are reduced. Consequently, SmaAt-Krige-GNet benefits from these maps in low-intensity rainfall scenarios, where larger-scale patterns dominate but struggle with sharp precipitation peaks, which require higher spatial resolution. Increasing the density of weather stations would directly improve the fidelity of the interpolated fields. A potential solution to this limitation would be a significant increase in the number of weather stations, which could improve the resolution and accuracy of the interpolated Kriging maps.

Fig. \ref{fig:input12} and Fig. \ref{fig:example-outputs} further clarifies the strengths and remaining limitations of the proposed architectures. Fig. \ref{fig:input12} illustrates four representative precipitation maps across different seasons. These inputs contains both stratiform and convective rainfall, providing a challenging test for models relying solely on reflectivity patterns. 2019-03-10, 2019-09-24, and 2019-12-14 correspond to stratiform precipitation events associated with large-scale frontal systems, characterized by broad rainfall distributions. In contrast, 2019-06-15 represents a convective summer case with sharp spatial gradients. Including both regimes enables a comprehensive evaluation of the models’ ability to handle diverse rainfall patterns.

\section{Conclusion}

This study introduced two novel deep learning models, SmaAt-fUsion and SmaAt-Krige-GNet, designed to improve precipitation nowcasting by integrating meteorological variables from weather stations alongside radar data. The experimental results demonstrate that both models have distinct advantages: SmaAt-Krige-GNet excels in low-intensity rainfall predictions by incorporating weather station data through Kriging-based spatial interpolation, while SmaAt-fUsion outperforms in high-intensity events by directly integrating station data into the network’s bottleneck. These findings highlight the potential of leveraging both discrete weather station data and spatially interpolated variables to improve deep learning-based precipitation forecasting. Future work includes extending the proposed fusion strategies to alternative architectures, additional multi-source meteorological datasets, and longer lead times to further assess their generality and robustness across diverse nowcasting settings.


\bibliographystyle{ieeetr}
\bibliography{references}

@inproceedings{stanczyk2021deep,
title = {Deep Graph Convolutional Networks for Wind Speed Prediction},
author={Sta\`{n}czyk, Tomasz and Mehrkanoon, Siamak},
booktitle = {European Symposium on Artificial Neural Networks, Computational Intelligence and Machine Learning (ESANN)},
pages = {147--152},
year = {2021},
}

@article{mehrkanoon2019deep,
  title={Deep shared representation learning for weather elements forecasting},
  author={Mehrkanoon, Siamak},
  journal={Knowledge-Based Systems},
  volume={179},
  pages={120--128},
  year={2019},
  publisher={Elsevier}
}

@article{ayzel2020rainnet,
  title={RainNet v1. 0: a convolutional neural network for radar-based precipitation nowcasting},
  author={Ayzel, Georgy and Scheffer, Tobias and Heistermann, Maik},
  journal={Geoscientific Model Development},
  volume={13},
  number={6},
  pages={2631--2644},
  year={2020},
  publisher={Copernicus GmbH}
}

@article{kaparakis2023wf,
  title={\uppercase{WF}-\uppercase{UN}et: Weather data fusion using 3d-unet for precipitation nowcasting},
  author={Kaparakis, Christos and Mehrkanoon, Siamak},
  journal={Procedia Computer Science},
  volume={222},
  pages={223--232},
  year={2023},
  publisher={Elsevier}
}

@article{precipitation_dataset,
  title={Extreme value modeling of areal rainfall from weather radar},
  author={Overeem, A and Buishand, TA and Holleman, I and Uijlenhoet, R},
  journal={Water Resources Research},
  volume={46},
  number={9},
  year={2010},
  publisher={Wiley Online Library}
}

@misc{pykrige,
  title={PyKrige: Kriging Toolkit for Python},
  author={Murphy, Benjamin S},
  year={2021},
  publisher={Version}
}

@article{Verworn2011,
  title={Spatial interpolation of hourly rainfall--effect of additional information, variogram inference and storm properties},
  author={Verworn, A and Haberlandt, Uwe},
  journal={Hydrology and Earth System Sciences},
  volume={15},
  number={2},
  pages={569--584},
  year={2011},
  publisher={Copernicus Publications G{\"o}ttingen, Germany}
}

@article{CSEGRecorderVariograms,
  title={The Variogram Basics: A visual introduction to one of the most useful geostatistical concepts},
  author={Brown, E},
  journal={CSEG},
  volume={47},
  pages={1--30},
  year={2022}
}

@article{Matheron1963,
  title={Principles of geostatistics},
  author={Matheron, Georges},
  journal={Economic geology},
  volume={58},
  number={8},
  pages={1246--1266},
  year={1963},
  publisher={Society of Economic Geologists}
}

@book{Isaaks1989,
  author = {Isaaks, E. H. and Srivastava, R. M.},
  title = {An Introduction to Applied Geostatistics},
  publisher = {Oxford University Press},
  year = {1989},
  isbn = {978-0195050134}
}

@article{mcc,
  title={The advantages of the Matthews correlation coefficient (MCC) over F1 score and accuracy in binary classification evaluation},
  author={Chicco, Davide and Jurman, Giuseppe},
  journal={BMC genomics},
  volume={21},
  pages={1--13},
  year={2020},
  publisher={Springer}
}

@article{trebing2021smaat,
  title={\uppercase{S}ma\uppercase{A}t-\uppercase{UN}et: Precipitation nowcasting using a small attention-UNet architecture},
  author={Trebing, Kevin and Sta\'{n}czyk, Tomasz and Mehrkanoon, Siamak},
  journal={Pattern Recognition Letters},
  volume={145},
  pages={178--186},
  year={2021},
  publisher={Elsevier}
}

@article{vatamany2025graph,
  title={Graph dual-stream convolutional attention fusion for precipitation nowcasting},
  author={Vatam{\'a}ny, L{\'o}r{\'a}nd and Mehrkanoon, Siamak},
  journal={Engineering Applications of Artificial Intelligence},
  volume={141},
  pages={109788},
  year={2025},
  publisher={Elsevier}
}

@article{ayzel2019all,
  title={All convolutional neural networks for radar-based precipitation nowcasting},
  author={Ayzel, G and Heistermann, M and Sorokin, A and Nikitin, O and Lukyanova, O},
  journal={Procedia Computer Science},
  volume={150},
  pages={186--192},
  year={2019},
  publisher={Elsevier}
}

@article{li2024quantitative,
  title={Quantitative Applications of Weather Satellite Data for Nowcasting: Progress and Challenges},
  author={Li, Jun and Zheng, Jing and Li, Bo and Min, Min and Liu, Yanan and Liu, Chian-Yi and Li, Zhenglong and Menzel, W Paul and Schmit, Timothy J and Cintineo, John L and others},
  journal={Journal of Meteorological Research},
  volume={38},
  number={3},
  pages={399--413},
  year={2024},
  publisher={Springer}
}

@article{wilson2010nowcasting,
  title={Nowcasting challenges during the Beijing Olympics: Successes, failures, and implications for future nowcasting systems},
  author={Wilson, James W and Feng, Yerong and Chen, Min and Roberts, Rita D},
  journal={Weather and Forecasting},
  volume={25},
  number={6},
  pages={1691--1714},
  year={2010},
  publisher={American Meteorological Society}
}

@article{reulen2024ga,
  title={\uppercase{GA}-\uppercase{S}ma\uppercase{A}t-\uppercase{GN}et: Generative adversarial small attention gnet for extreme precipitation nowcasting},
  author={Reulen, Eloy and Shi, Jie and Mehrkanoon, Siamak},
  journal={Knowledge-Based Systems},
  volume={305},
  pages={112612},
  year={2024},
  publisher={Elsevier}
}

@article{bauer2015quiet,
  title={The quiet revolution of numerical weather prediction},
  author={Bauer, Peter and Thorpe, Alan and Brunet, Gilbert},
  journal={Nature},
  volume={525},
  number={7567},
  pages={47--55},
  year={2015},
  publisher={Nature Publishing Group UK London}
}

@article{ham2023anthropogenic,
  title={Anthropogenic fingerprints in daily precipitation revealed by deep learning},
  author={Ham, Yoo-Geun and Kim, Jeong-Hwan and Min, Seung-Ki and Kim, Daehyun and Li, Tim and Timmermann, Axel and Stuecker, Malte F},
  journal={Nature},
  volume={622},
  number={7982},
  pages={301--307},
  year={2023},
  publisher={Nature Publishing Group UK London}
}

@article{allen2025end,
  title={End-to-end data-driven weather prediction},
  author={Allen, Anna and Markou, Stratis and Tebbutt, Will and Requeima, James and Bruinsma, Wessel P and Andersson, Tom R and Herzog, Michael and Lane, Nicholas D and Chantry, Matthew and Hosking, J Scott and others},
  journal={Nature},
  volume={641},
  number={8065},
  pages={1172--1179},
  year={2025},
  publisher={Nature Publishing Group}
}

@book{wilks2011statistical,
  title={Statistical Methods in the Atmospheric Sciences},
  author={Wilks, Daniel S.},
  year={2011},
  publisher={Academic Press}
}

@article{jiang2024hybrid,
  title={Hybrid multilayer perceptron and convolutional neural network model to predict extreme regional precipitation dominated by the large-scale atmospheric circulation},
  author={Jiang, Qin and Cioffi, Francesco and Li, Weiyue and Tan, Jinkai and Pan, Xiaoduo and Li, Xin},
  journal={Atmospheric Research},
  volume={304},
  pages={107362},
  year={2024},
  publisher={Elsevier}
}

@article{wang2025enhancing,
  title={Enhancing Precipitation Nowcasting Through Dual-Attention RNN: Integrating Satellite Infrared and Radar VIL Data},
  author={Wang, Hao and Yang, Rong and He, Jianxin and Zeng, Qiangyu and Xiong, Taisong and Liu, Zhihao and Jin, Hongfei},
  journal={Remote Sensing},
  volume={17},
  number={2},
  pages={238},
  year={2025},
  publisher={MDPI}
}

@article{cao2025enhancing,
  title={Enhancing nowcasting with multi-resolution inputs using deep learning: Exploring model decision mechanisms},
  author={Cao, Yuan and Chen, Lei and Wu, Junjing and Feng, Jie},
  journal={Geophysical Research Letters},
  volume={52},
  number={4},
  pages={e2024GL113699},
  year={2025},
  publisher={Wiley Online Library}
}

\end{document}